\documentclass{article} % For LaTeX2e
\usepackage{nips14submit_e,times}
\usepackage{hyperref}
\usepackage{rotating}
\usepackage{url}
\usepackage{tabularx}
\usepackage[tight,footnotesize]{subfigure}
\usepackage{amsmath}
\usepackage{wrapfig}
\title{Semantic Graph for Zero-Shot Learning}

\author{
Zhen-Yong Fu, Tao Xiang and Shaogang Gong\\
School of Electronic Engineering and Computer Science\\
Queen Mary University of London, UK\\
\texttt{z.fu@qmul.ac.uk}
}

\nipsfinalcopy % Uncomment for camera-ready version

\begin{document}

\maketitle
%------------------------------------------------------------------------------------------------
\begin{abstract}
Zero-shot learning aims to classify visual objects without any training data
via knowledge transfer between seen and unseen classes. This is typically
achieved by exploring a semantic embedding space where the seen and unseen
classes can be related. Previous works differ in what embedding space is used
and how different classes and a test image can be related. In this paper, we
utilize the annotation-free semantic word space for the former and focus on
solving the latter issue of modeling relatedness. Specifically, in contrast to
previous work which ignores the semantic relationships between seen classes
and focus merely on those between seen and unseen classes, in this paper a
novel approach based on a semantic graph is proposed to represent the
relationships between all the seen and unseen class in a semantic word
space. Based on this semantic graph, we design a special absorbing Markov
chain process, in which each unseen class is viewed as an absorbing
state. After incorporating one test image into the semantic graph, the
absorbing probabilities from the test data to each unseen class can be
effectively computed; and zero-shot classification can be achieved by finding
the class label with the highest absorbing probability. The proposed model has
a closed-form solution which is linear with respect to the number of test
images. We demonstrate the effectiveness and computational efficiency of the
proposed method over the state-of-the-arts on the AwA (animals with
attributes) dataset.
\end{abstract}

%------------------------------------------------------------------------------------------------
\section{Introduction}
\label{sec:introduction}
Zero-shot learning (ZSL) for visual classification has received increasing
attentions recently
\cite{lampert2009learning,rohrbach2010helps,rohrbach2011evaluating,frome2013devise,norouzi2013zero}. This
is because although virtually unlimited images are available via social media
sharing websites such as Flickr, there are still not enough annotated images
for building a visual classification model for a large number of visual
classes. ZSL aims to imitate human's ability to recognize a new class without
even seeing any instance. A human has that ability because he/she is able to
make connections between an unseen class with the seen classes based on its
semantic description. Similarly a zero-shot learning method for visual
classification relies on the existence of a labeled training set of seen
classes and the knowledge about how each unseen class is semantically related
to the seen classes.

An unseen class can be related to a seen class by representing both in a
semantic embedding space \cite{frome2013devise}. Existing ZSL methods can be
categorized by the different embedding spaces deployed. Early works are
dominated by semantic attribute based approaches. Visual classes are embedded
in to an attribute space by defining an attribute ontology and annotating a
binary attribute vector for each class. The similarity between different
classes can thus be measured by how many attributes are shared. However, both
the ontology and attribute vector for each class need to be manually defined
with the latter may have to be annotated at the instance level due to large
intra-class variations. This gives poor scalability to these attribute-based
approaches \cite{norouzi2013zero}. Alternatively, recently embedding based on
semantic word space started to gain popularity
\cite{frome2013devise,norouzi2013zero}. Learned from a large language corpus,
this embedding space is `free' and applicable to any visual classes
\cite{mikolov2013distributed, mikolov2013efficient}. It thus has much better
scalability and is the embedding space adopted in this paper. 

After choosing an embedding space, the remaining problem for a ZSL approach is
to measure the similarity between a test data with each unseen class so that
(zero-shot) classification can be performed. Since there is no training data
for the unseen classes, such a similarity obviously cannot be computed
directly and the training data from the seen classes need to be explored to
compute the similarity indirectly. Again, two options are available. In the
first option, the seen class data are used to learn a mapping function to map
a low-level feature representation of a training image to the semantic
space. Such a mapping function is then employed to map a test image belonging
a unseen class to the same space where similarity between the data and a class
embedding vector can be computed for classification
\cite{frome2013devise}. However, this approach has an intrinsic limitation --
the mapping function learned from the seen class may not be suitable for the
unseen classes due to the domain shift problem. Rectifying this problem by
adapting the mapping function to the unseen classes is also hard as no labeled
data is available for those classes. The second option is to avoid the need
for mapping a test image into the semantic embedding space. The training data
is used in a different way -- instead of learning a mapping function from the
low-level feature to the semantic embedding space, a n-way probabilistic
classifier is learned in the visual feature space. The embedding space is used
purely for computing the semantic relatedness \cite{rohrbach2010helps} between
the seen and unseen classes. This semantic relatedness based approach
alleviates the domain shift problem and has been empirically shown to be
superior to the direct mapping based approach
\cite{lampert2013attribute,norouzi2013zero}. It is thus the focus of this
paper.

\begin{figure}
  \begin{center}
    \subfigure[Bipartite graph]{\label{fig:subfig:direct_similarity_based}\includegraphics[width=0.22\linewidth]{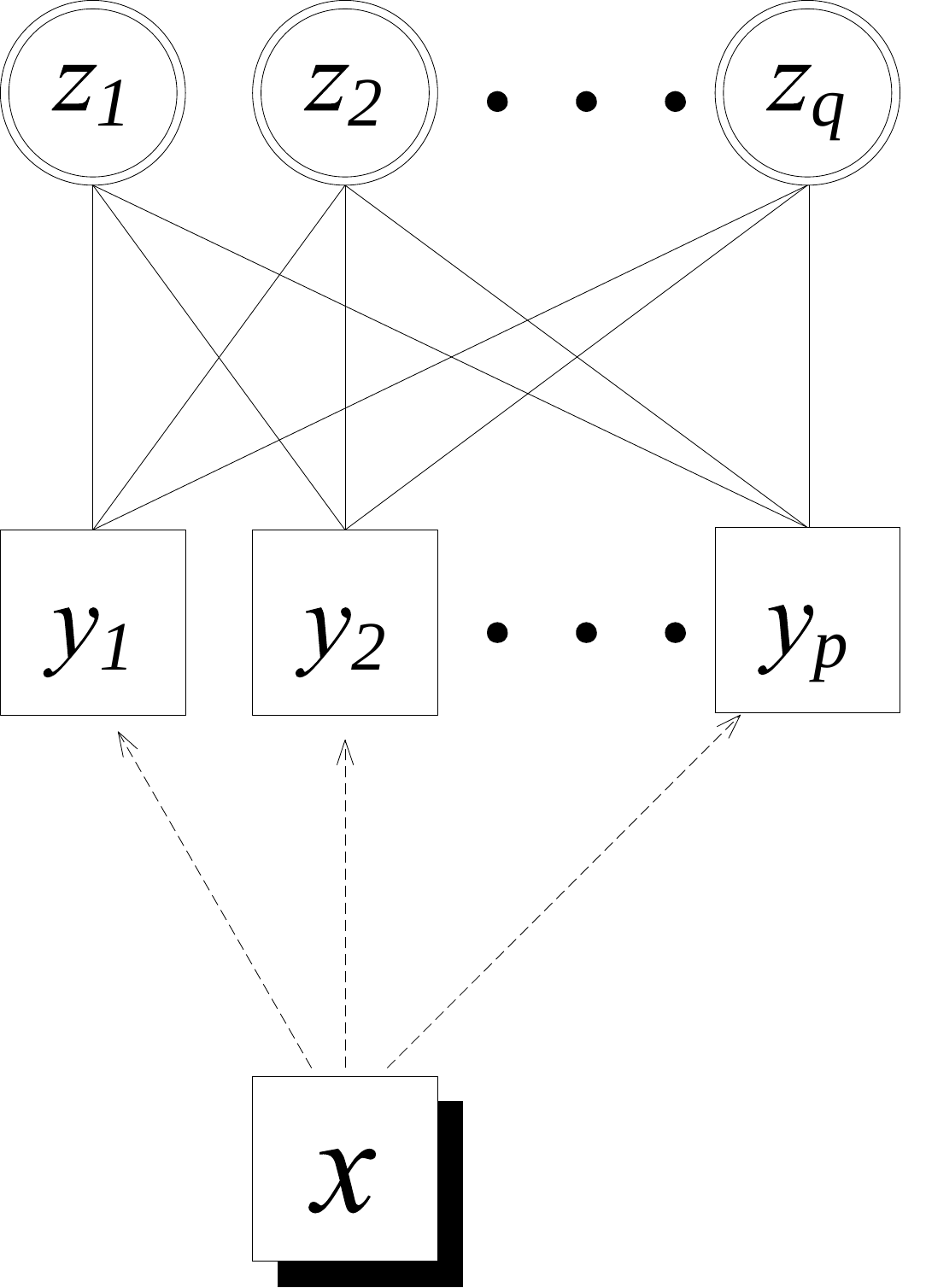}}\hspace{0.09\linewidth}
    \subfigure[Semantic graph]{\label{fig:subfig:amp_based}\includegraphics[width=0.68\linewidth]{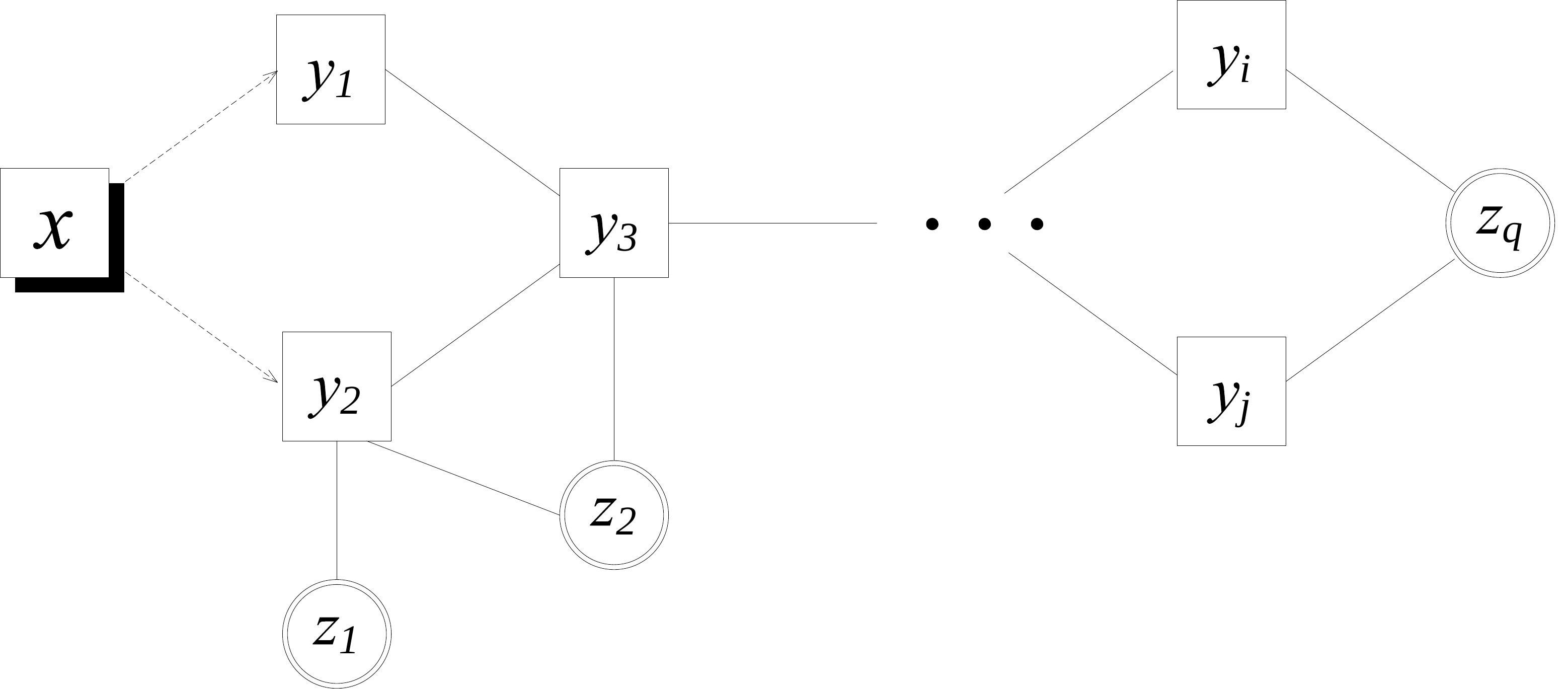}}
  \end{center}
  \caption{Bipartite graph \cite{rohrbach2010helps} vs. the proposed semantic graph for zero-shot object classification. An unseen data point is denoted as $x$; the $i$-th seen and unseen classes are denoted as $y_i$ and $z_i$ respectively. }
  \label{fig:illustration}
\end{figure}

In this paper, a novel semantic graph based approach is proposed to model the
relatedness between seen and unseen classes. In previous work
\cite{rohrbach2010helps}, the relatedness between seen and unseen classes is
modeled with a bipartite graph.  As shown in
Fig.~\ref{fig:subfig:direct_similarity_based}, in such a graph the
relatedness between each unseen class and each seen class is modeled directly
in a flat structure, while the relatedness between the seen classes is
ignored. This can be viewed as an `one step' exploration in the bipartite
graph. In contrast, in this paper, we extend the modeling for semantic
relationships from the flat structure to a hierarchical structure and perform
a multiple-step exploration. As shown in Fig.~\ref{fig:subfig:amp_based}, in
our approach, seen and unseen classes will form a semantic graph, in which
each seen or unseen class corresponds to a graph node. The semantic graph is
constructed as a $k$-nearest-neighbor (nn)  graph. It should be noted that on
a semantic graph, the relatedness between seen classes is modeled explicitly;
in addition, each unseen class can only connect with seen classes and there is
no direct connection among unseen classes. In this way the relatedness between
different seen classes are also exploited, making the similarity measure
between a test image and each unseen class more robust. Furthermore, compared
to the bipartite graph, the $k$-nn semantic graph can be computed  more
efficiently. For example, for $p$ seen classes and $q$ unseen classes, the
bipartite graph needs to store $O(pq)$ parameters (the weights on the graph
edges), while the $k$-nn semantic graph only needs to store $O(k(p+q))$
parameters.

More specifically, for a test image $x$, to perform the zero-shot learning, we
connect it to the seen class nodes, that is, we incorporate $x$ into the
semantic graph. Different with the bipartite graph-based method
\cite{rohrbach2010helps}, it is possible that there is no direct connection
between the real target unseen class and the seen classes connected by the
test image on the semantic graph. Consequently, we have to design a new
approach so that if the test image and an unseen class are connected with
shorter paths on the semantic graph, the test image should have higher
probability to be labeled as that unseen class. For example, in
Fig.~\ref{fig:subfig:amp_based}, the test image $x$ should have higher
probabilities to be classified to unseen class $z_1$ or $z_2$ than $z_q$. To
this end, we define a special absorbing Markov chain process on the semantic
graph. We view each unseen class node as the absorbing state. Thus, each path
that starts from $x$ and terminates at one unseen class will not include other
unseen classes. The inner nodes of such kind of paths only include the seen
class nodes. The seen class nodes can thus be viewed as the bridge nodes that
connect the test image and the unseen classes. The absorbing probabilities
from the test image to each unseen class can be effectively computed. Given
the predicted absorbing probabilities, we perform zero-shot learning by
finding the class label with highest absorbing probability. Moreover, we show
that the proposed method has a closed-form solution which is  linear with
respect to the number of test images.

The main contributions of this work are as follows. First, we propose to use
the $k$-nearest-neighbor semantic graph to model the relatedness among seen
and unseen classes. This makes the similarity measure between a test image and
unseen classes more robust, and as the number of visual categories increases,
compared to bipartite graph, our $k$-nn semantic graph will be more
efficient. Second, we design a special absorbing Markov chain process on the
semantic graph and show how to effectively compute the absorbing probabilities
from one test image to each of unseen classes. Third, after stacking the
absorbing probabilities for each test image together, we provide a zero-shot
learning algorithm that has a closed-form solution and is a linear  with
respect to the number of test images.

The remainder of this paper is organized as follows. After a review of
previous work (Section \ref{sec:related_work}), we first introduce our
approach (Section \ref{sec:approach}) and then give experimental results
(Section \ref{sec:experiments}). The paper concludes  in Section
\ref{sec:conclusion}. 

%------------------------------------------------------------------------------------------------
\section{Previous Work}
\label{sec:related_work}
%P1, embedding spaces for ZSL. People used attributes but it is
%unsalable. Then people use word vectors. Our work also use word vector but in
%a very different way. 

\textbf{Semantic embedding for ZSL}.  In most earlier works on zero-shot
learning, semantic attributes are employed as the embedding space for
knowledge transfer
\cite{lampert2009learning,palatucci2009zero,farhadi2009describing,farhadi2010attribute,akata2013label}. Most
existing studies  assume that an exhaustive ontology of attributes has been
manually specified at either the class or instance level
\cite{lampert2009learning,scheirer2012multi}. However, annotating attributes
scales poorly as ontologies tend to be domain specific. For example, birds and
trees have very different set of attributes. Some works proposed to
automatically learn discriminative visual attributes from data
\cite{ferrari2007learning,farhadi2009describing}. But this sacrifices the
name-ability of the embedding space as the discovered attributes may not be
semantically meaningful.  To overcome this problem, semantic representations
that do not rely on an explicit attribute ontology have been proposed
\cite{rohrbach2010helps,rohrbach2011evaluating}. In particular, recently
semantic word space has been investigated
\cite{Oquab14,socher2013zero,frome2013devise}.  A word space is extracted from
linguistic knowledge bases e.g. WordNet or Wikipedia by natural language
processing models. Instead of manually defining an attribute prototype, a
novel target class' textual name can be projected into this space and then
used as the prototype for zero-shot learning. Typically learned from a large
corpus covering all English words and bi-grams, this word space can be used
for any visual classes without the need for any manual annotation. It is thus
much scalable than an attribute embedding space for ZSL. In this work, we
choose the word space for its scalability, but our method differs
significantly from \cite{Oquab14,socher2013zero,frome2013devise} in how the
embedding space is used for knowledge transfer and we show superior
performance experimentally (see Section \ref{sec:experiments}).

%P2, How knowledge is transferred from the seen classes. Most work transfer by learn a direct mapping, DAP or its variant. but there is domain shift problem. Few works learn classifier and use the classifier score to indirectly link the test image into the semantic word space. But focus is only on the relatedness between seen and unseen classes, the intrinsic manifold structure of the seen classes are exploited making them vulnerable to noise. 

\textbf{Knowledge transfer via an embedding space}. Given an embedding space,
existing approaches differ significantly in how the knowledge is transferred from
a labeled training set containing seen classes. Most existing approaches,
such as direct attribute prediction (DAP) \cite{lampert2009learning} or its
variants \cite{frome2013devise} take a directly mapping based
strategy. Specifically, the training data set is used to learn a mapping
function from the low-level feature space to the semantic embedding
space. Once learned, the same mapping function is used to map a test image in
to the same space where the similarity between the test image to each unseen
class semantic vector or prototype can be measured
\cite{frome2013devise}. This strategy however suffers from the mapping domain
shift problem mentioned earlier. Alternatively, a semantic relatedness based
strategy can be adopted. This involves learning a n-way probabilistic
classifier in the low-level feature space for the training seen
classes. Given a test image, the probabilities produced by this classifier for
each seen class indicate the visual similarity or relatedness between the
test image and the seen classes. This relatedness is then compared with the
semantic relatedness between each unseen class and the same seen classes. The
test image is then classified according to how the visual similarity and
semantic similarity agree. One representative approach following this strategy
is Indirect Attribute Prediction (IAP) \cite{lampert2013attribute}.  It has
also been shown that the semantic relatedness does not necessarily come from a
semantic embedding space, e.g.~ it can be computed from hit counts from an
image search engine \cite{rohrbach2010helps}.  This indirect semantic
relatedness based strategy can be potentially advantageous over the direct
mapping based one, as verified by the results in
\cite{lampert2013attribute,norouzi2013zero}. However, as we analyzed earlier,
the existing approaches based on semantic relatedness employ a flat bipartite
graph and ignore the important inter-seen-class relatedness. In this work we
develop a novel semantic graph based zero-shot learning method and show its
advantages over the bipartite graph based methods on both classification
performance and computational efficiency.

%{\bf TX to Ian: }P3, ideally we want to make connections to other graph or manifold learn based approach beyond the scope of ZSL. If short of time or space, we can leave it out.

%------------------------------------------------------------------------------------------------
\section{Approach}
\label{sec:approach}
%------------------------------------------------------------------------------------------------
\subsection{Problem Definition}
\label{sec:problem}
Let $\mathcal{Y}=\{y_1,\mathellipsis,y_p\}$ denote the seen classes set and
$\mathcal{Z}=\{z_1,\mathellipsis,z_q\}$ denote the unseen classes set.
Given a training dataset $X_{\mathcal{Y}}$ labeled as $y_j\in\mathcal{Y}$, the goal
of zero-shot learning is to learn a classifier $f: X\rightarrow\mathcal{Z}$
even if there is no training data labeled as $z_j\in\mathcal{Z}$.

Taking a semantic relatedness strategy for knowledge transfer, we first
utilize the training dataset $X_{\mathcal{Y}}$ to learn a classifier for the
seen classes $\mathcal{Y}$. In this paper, we use the support vector machine
(SVM) as the classifier for seen classes. For a test image $x_i\notin
X_{\mathcal{Y}}$, the SVM classifier can provide an estimate of the posterior
probability $p(y_j|x_i)$ of image $x_i$ belonging to seen class $y_j$. Let
$t_{i,\cdot}=[t_{ij}]_{1\times p}$ be a row vector with $p$ elements, in which
each element $t_{ij}=p(y_j|x_i)$. For a whole test dataset with $n$ images, we
will have the matrix $T=[t_{ij}]_{n\times p}$, in which each row corresponds
to a test image $x_i$. $T$ stores the relationship between the test images and
the seen classes. It should be noted that although in this work, this
relationship is measured by the posterior probability $p(y_j|x_i)$, other ways
of computing the relationship between test images and the seen classes can
also be adopted.

Our objective is to perform zero-shot learning through modeling the
relationship between seen classes $y_1,\mathellipsis,y_p$ and unseen classes
$z_1,\mathellipsis,z_q$. In this paper, we propose to use semantic graph to
model the relationship among classes.

%------------------------------------------------------------------------------------------------
\subsection{Semantic Graph}
\label{sec:semantic_graph}

For measuring the relationship between two classes, we employ the word vector
representation from the linguistic research
\cite{mikolov2013efficient,mikolov2013distributed} and use the $cosine$
similarity of their word vectors as the similarity measurement of the two
classes.

Furthermore, a semantic graph is constructed as a $k$-nearest-neighbor
graph. In the semantic graph, each class (regardless if it is a seen or unseen
class) will have a corresponding  graph node which is connected with its $k$
most similar (semantically related) other classes. The edge weight $w_{ij}$ of
the semantic graph is the $cosine$ similarity between two end node of this
edge. More details about the semantic graph construction can be found in
Section \ref{sec:exp_setup}. After constructing the semantic graph, the graph
structure will be fixed in the next steps of the pipeline. 

We then define a special absorbing Markov chain process on the semantic graph,
in which each unseen class node is viewed as an \emph{absorbing} state and
each seen class node is viewed as \emph{transient} state. The transition
probability from class node $i$ to class node $j$ is
$p_{ij}=w_{ij}/\sum_j{w_{ij}}$, i.e. the normalized similarity. The absorbing
state means that for each unseen class node $i$, we have $p_{ii}=1$ and
$p_{ij}=0$ for $i\neq j$. It should be noted that since all of the unseen
class nodes are absorbing states, there will have no \emph{direct} connection
between two unseen class nodes. In other words, the unseen classes will be
connected through the seen classes.

We re-number the class nodes (states in Markov process) so that the seen class
nodes (transient states) come first. Then, the transition matrix $P$ of the above
absorbing Markov chain process will have the following canonical form:
\begin{equation}
  \label{P_matrix}
  P=
  \left(
    \begin{array}{c|c}
      Q_{p\times p} & R_{p\times q}\\\hline
      \mathbf{0}_{q\times p} & \mathbf{\mathrm{I}}_{q\times q}
    \end{array}
  \right).
\end{equation}In El.~\ref{P_matrix}, $Q_{p\times p}$ describes the probability
of transitioning from a transient state (seen class) to another and
$R_{p\times q}$ describes the probability of transitioning from a transient
state (seen class) to an absorbing state (unseen class). In addition,
$\mathbf{0}_{q\times p}$ and the identity matrix $\mathbf{\mathrm{I}}_{q\times
  q}$ mean that the absorbing Markov chain process cannot leave the absorbing
states once it arrives.

\begin{figure}[t]
  \begin{center}
    \includegraphics[width=0.8\linewidth]{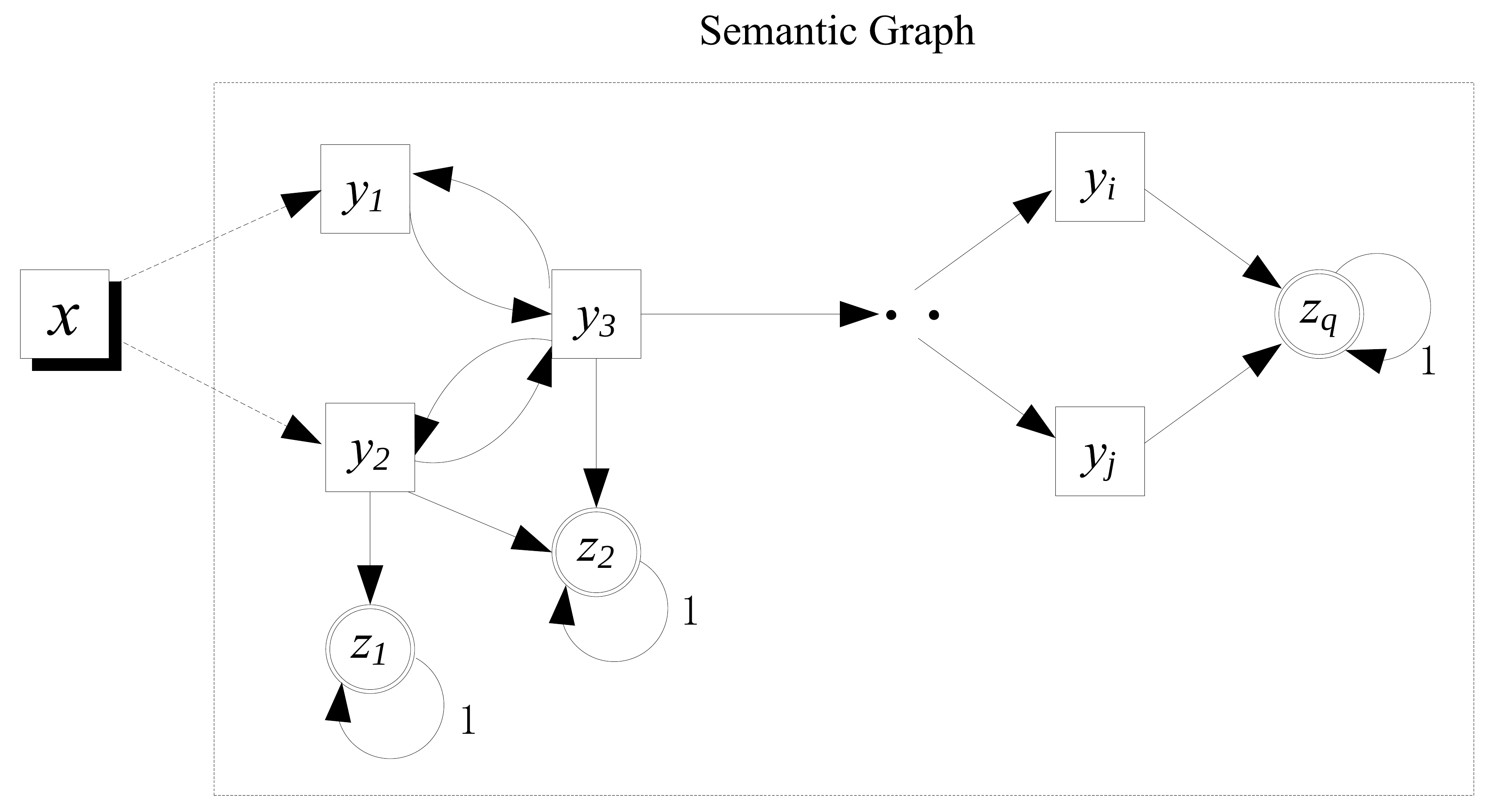}
  \end{center}
  \caption{After incorporating the test image into the semantic graph,
    zero-shot learning can be viewed as an absorbing Markov chain process on
    semantic graph.}
  \label{fig:amp_process}
\end{figure}

%------------------------------------------------------------------------------------------------
\subsection{Zero-shot Learning}
\label{sec:zsl}
For zero-shot learning, i.e. predicting the label of an unseen image $x_i$, we
first need to incorporate $x_i$ into the semantic graph. And then we will
apply an extended absorbing Markov chain process, in which the test image
$x_i$ is involved, to perform the zero-shot learning.

In order to introduce a test image $x_i$ into the semantic graph, it is
connected with some seen class nodes \footnote{Obviously it cannot be
  connected to the unseen class nodes directly as we are not mapping $x_i$ in
  to the same semantic space.}. The nodes selected for connection is
determined by the posterior probability $p(y_j|x_i)$ of image $x_i$ belonging
to seen class $y_j$. Specifically, the node representing image $x_i$ is
connected to the seen classes with the highest posterior probability,
i.e.~most visually similar. Note that for $x_i$, there will have no stepping
in probabilities and the Markov process can only step out from $x_i$ to other
seen class nodes. The stepping out probabilities from $x_i$ to seen class
nodes are $t_{i,\cdot}$, which are the posterior probability computed using
the seen class classifier as described in Section \ref{sec:problem}.  $x_i$ is
thus incorporated into the semantic graph as a transient state. The transition
matrix $\tilde{P}$ of the extended absorbing Markov chain process have the
following canonical form:
\begin{equation}
  \label{extended_P_matrix}
  \tilde{P}=
  \left(
    \begin{array}{cc|c}
      Q_{p\times p} & \mathbf{0}_{p\times 1} & R_{p\times q}\\
      (t_{i,\cdot})_{1\times p} & 0_{1\times 1} & \mathbf{0}_{1\times q}\\\hline
      \multicolumn{2}{c|}{\mathbf{0}_{q\times (p+1)}} & \mathbf{\mathrm{I}}_{q\times q}
    \end{array}
  \right).
\end{equation}In the meanwhile, the extended transition matrix within all
transient states, including all seen class nodes and one extra test image node
$x_i$, are written as
\begin{equation}
  \label{Q_prime}
  \tilde{Q}_{(p+1)\times(p+1)}=
  \left(
    \begin{array}{cc}
      Q_{p\times p} & \mathbf{0}_{p\times 1}\\
      (t_{i,\cdot})_{1\times p} & 0_{1\times 1}
    \end{array}
  \right),
\end{equation}and the extended transition matrix between transient states and
absorbing states should be
\begin{equation}
  \label{R_prime}
  \tilde{R}_{(p+1)\times q}=
  \left(
    \begin{array}{c}
      R_{p\times q}\\
      \mathbf{0}_{1\times q}
    \end{array}
  \right).
\end{equation}

In the extended semantic graph, it is obvious that if there are many short
paths that connect the test image node $x_i$ and one unseen class node,
e.g. $z_j$, the absorbing Markov chain process that starts from $x_i$ will
have a high probability to be absorbed at $z_j$. Thus, the probability that
$x_i$ is labeled as $z_j$ should be high. This is a cumulative process and can
be reflected by the absorbing probabilities from $x_i$ to all unseen class
nodes.

The absorbing probability $b_{ij}$ is the probability that the absorbing
Markov chain will be absorbed in the absorbing state $s_j$ if it starts from
the transient state $s_i$. The absorbing probability matrix
$\tilde{B}=[b_{ij}]_{(p+1)\times q}$ can be computed as follows:
\begin{equation}
  \label{B_prime}
  \tilde{B} = \tilde{N}\times\tilde{R},
\end{equation}in which $\tilde{N}$ is the fundamental matrix of the extended
absorbing Markov chain process and is defined as follows:
\begin{equation}
  \label{N_prime}
  \tilde{N}_{(p+1)\times (p+1)}=(\mathbf{\mathrm{I}}-\tilde{Q})^{-1}=
  \left(
    \begin{array}{cc}
      \mathbf{\mathrm{I}}_{p\times p}-Q_{p\times p} & \mathbf{0}_{p\times 1}\\
      -(t_{i,\cdot})_{1\times p} & 1
    \end{array}
  \right)^{-1}.
\end{equation}

We use the following block matrix inversion formula to compute
$\tilde{N}$. 
\begin{equation*}
  \left(
    \begin{array}{cc}
      A & B\\
      C & D
    \end{array}
  \right)^{-1}
  =
  \left(
    \begin{array}{cc}
      (A-BD^{-1}C)^{-1} & -(A-BD^{-1}C)^{-1}BD^{-1}\\
      -(D-CA^{-1}B)^{-1}CA^{-1} & (D-CA^{-1}B)^{-1}
    \end{array}
  \right).
\end{equation*}Since we only care about the absorbing probabilities that the
absorbing chain process starts from the test image node $x_i$, we only need to
compute the last row of $\tilde{B}$, i.e. $\tilde{B}_{p+1,\cdot}$ for $x_i$
($x_i$ corresponds to the last transient state in the extended canonical form
in Eq.~\ref{extended_P_matrix}). In particular, we can apply the above block
matrix inversion formula to compute the last row of $\tilde{N}$ as
\begin{equation}
  \tilde{N}_{(p+1),\cdot}=
  \left(
    \begin{array}{cc}
      (t_{i,\cdot})(I-Q)^{-1}, & 1
    \end{array}
  \right)_{1\times(p+1)}
\end{equation}and then we may further compute $\tilde{B}_{p+1,\cdot}$ as
\begin{equation}
  \tilde{B}_{p+1,\cdot}=(\tilde{N}_{(p+1),\cdot})\times \tilde{R}=(t_{i,\cdot})(I-Q)^{-1}R.
\end{equation}

For the whole test dataset with $n$ images, we use a matrix $S_{n\times q}$ to
store the computed absorbing probabilities, in which the $i$-th row
$S_{i,\cdot}$ of $S$ equals to the absorbing probabilities of $x_i$. If we
stack the results of all test images together, we will get the final matrix
$S$ as follows,
\begin{equation}
  \label{S_matrix}
  S = T(I-Q)^{-1}R.
\end{equation}
In Eq.~\ref{S_matrix}, $T$ is a $n\times p$ matrix and $(I-Q)^{-1}R$ is a
$p\times q$ matrix that is only related to the semantic graph structure and
can be pre-computed. The only dimension variable in
Eq.~\ref{S_matrix} is the number of test images $n$. Therefore, our method is
linearly with respect to the number of test images.

Finally, for the test image $x_i$, we assign it to the unseen label that has the
maximum absorbing probability when the absorbing chain starts from $x_i$. That is,
\begin{equation}
f(x_i)=\arg\max_jS_{i,j}
\end{equation}

It should be noted that in our formulation, we consider all the paths in the
semantic graph, i.e. the whole structure of the semantic graph. Therefore, our
method is more stable compared to direct similarity-based zero-shot learning,
in the sense of being less sensitive to the number of connections to the seen
classes for each test image, and the imperfect seen class classifier causing
noise in the posterior probability computed. This is verified by the
experimental results in Section \ref{sec:results}.

%------------------------------------------------------------------------------------------------
\section{Experiments}
\label{sec:experiments}
\subsection{Experimental Setup}
\label{sec:exp_setup}
\begin{table}[t]
  \begin{center}
    \begin{tabular}{l | c c c c c c c c c c | c || c}
      & \multicolumn{11}{c||}{Area under ROC curve (AUC) in \%} & \\
      Methods & \begin{turn}{90}chimpanzee\end{turn} & \begin{turn}{90}giant
        panda\end{turn} & \begin{turn}{90}leopard\end{turn}
      & \begin{turn}{90}Persian cat\end{turn} & \begin{turn}{90}pig\end{turn}
      & \begin{turn}{90}hippopotamus\end{turn} & \begin{turn}{90}humpback
        whale\end{turn} & \begin{turn}{90}raccoon\end{turn}
      & \begin{turn}{90}rat\end{turn} & \begin{turn}{90}seal\end{turn}
      & \begin{turn}{90}mean AUC (in \%)\end{turn} & \begin{turn}{90}mean
        accuracy (in \%)\end{turn} \\
      % \hline
      % \hline
      % DAP (decaf) \cite{lampert2013attribute} & & & & & & & & & & & 81.4 &
      % 41.4\\
      % IAP (decaf) \cite{lampert2013attribute} & & & & & & & & & & & 80.0 &
      % 42.2\\
      \hline
      \hline
      direct-similarity \cite{rohrbach2010helps} & 76 &	\textbf{73} & 84 & 78
      & \textbf{76} & \textbf{78} & 98 & 73 & 82 & \textbf{77} & 79.7 & 39.8\\
      ConSE \cite{norouzi2013zero} & 76	& 49 & \textbf{85} & 71 & 71 & 65 & \textbf{99} & 72 &
      81 & 72 & 74.1 & 35.1\\
      SVR+NN & 86 & 63 & 80 & \textbf{87} & 73 & 75 & \textbf{99} & 75 &
      \textbf{87} & 74 & \textbf{80.0} & 33.4\\
      Our method & \textbf{88} & 58 & 77 & \textbf{87} & 71 & \textbf{78} & \textbf{99}
      & \textbf{82} & \textbf{87} & 72 & 79.8 & \textbf{43.1}\\ 
      \hline
    \end{tabular}
  \end{center}
  \caption{Zero-shot classification results on the AwA dataset
    \cite{lampert2009learning}. The best results per table column are indicated in bold.}
  \label{table:main_result}
\end{table}

\textbf{Dataset.}
We utilize the AwA (animals with attributes) dataset
\cite{lampert2013attribute} to evaluate the performance of the proposed
zero-shot learning method. AwA provides 50 classes of animals (30475 images)
and 85 associated class-level attributes (such as furry, and hasClaws). In
this work, attributes are not used unless otherwise stated. AwA also provides
a defined source/target split for zero-shot learning with 10 classes and 6180
images held out.

\textbf{Competitors.} Our method is compared against three alternatives. The
first two are the most related, namely Rohrbach et al.'s direct
similarity-based ZSL (DS-based) \cite{rohrbach2010helps} and Norouzi et al.'s
convex semantic embedding ZSL (ConSE) \cite{norouzi2013zero}. Both methods
take a semantic relatedness strategy and learn a n-way probabilistic
classifier for the seen classes. In DS-based zero-shot learning, the semantic
relatedness  among categories are modeled as a bipartite graph. ConSE will
choose the top $K$ similar seen classes for a test image using the trained
classifier, and then use the prototypes of the seen classes in the word space
to form a new word vector for the test image. Zero-shot learning is performed
by finding the most similar unseen prototype in the word space. In addition,
we also apply the support vector regression to train a mapping from visual
space to word space and after mapping each test image into the word space, the
nearest-neighbor classifier is used to perform zero-shot learning. We call
this direct mapping based method SVR+NN. This method differs from the other
two and ours in that it uses the training data of seen classes to learn a
mapping function rather than a classifier. Apart from these three, we also
compare with the published results using attribute space rather than the
semantic word space.

\textbf{Settings.} We first exploit the word space representation
\cite{mikolov2013distributed,mikolov2013efficient} to transform each AwA seen
or unseen class name to a vector in the word space. For the word space, we
train the skip-gram text model on a corpus of 4.6M Wikipedia documents to form
a 1000-D word space. Since the \emph{seal} unseen class name of AwA has many
meanings in English, not just the animal seal, we choose seven concrete seal
species from the `seals-world'
website\footnote{http://www.seals-world.com/seal-species/}, that is, leopard
seal, harp seal, harbour seal, gray seal, elephant seal, weddell seal and monk
seal, to generate word vector for unseen \emph{seal} class. We use the decaf
feature \cite{donahue2013decaf} that is provided at the AwA
website\footnote{http://attributes.kyb.tuebingen.mpg.de/} and apply the libsvm
\cite{chang2011libsvm} to train a linear kernel SVM with probability estimates
output. All other parameters in libsvm are set to the default value. For
training SVR mapping, we apply the liblinear toolbox \cite{fan2008liblinear}
and set the parameter $C=10$. For semantic graph construction, we choose
different $k$ for seen classes and unseen classes when searching for the
$k$-nearest-neighbors. That is, we first construct a subgraph with seen
classes, in which we choose $k=2$. For the similarity matrix $W$ of the seen
subgraph, we set $W=(W+W^T)/2$ to ensure that it is symmetric for the seen
classes. For each unseen class, we connect it with top $k=4$ similar seen
classes according to the cosine similarity in the word space. This will ensure
that each unseen class is connected into the seen subgraph and there is no
isolated unseen class node on semantic graph. The code of our method can be
found at \footnote{https://sites.google.com/site/zhenyongfu10/}.

\subsection{Results}
\label{sec:results}
Table~\ref{table:main_result} compare the zero-shot classification performance
measure by  area under ROC curve (AUC) scores for the ten individual test
classes and their average. The last column in Table~\ref{table:main_result}
gives the corresponding average multi-class classification accuracies. In
DS-ZSL, ConSE and our method, each test image will be connected with $K=5$
seen classes. From Table \ref{table:main_result}, we can see that the proposed
semantic graph based method can achieve the best AUC results at six individual
test classes and the best average multi-class classification accuracy. As for
the average AUC on the ten test classes, the results of direct
similarity-based method, SVR+NN and our method are almost the same. SVR+NN
achieves the best average AUC result, but its average multi-class
classification accuracy is the lowest.

\textbf{Comparison with attribute-based ZSL.} We also compare our result with
the state-of-the-art results of attribute-based ZSL methods, including Lampert
et al.'s DAP and IAP \cite{lampert2013attribute} and Akata et
al.'s
% \begin{wraptable}{r}{0.5\textwidth}
%   \caption{Comparison with the state-of-the-art attribute-based zero-shot
%     learning on AwA.}
%   \label{table:comparison_with_attribute}
%   \begin{tabular}{c|c}\hline
%       Approach & mean accuracy (in \%) \\ \hline
%       DAP & 40.5(\cite{lampert2009learning}) / 41.4(\cite{lampert2013attribute}) \\
%       IAP & 27.8(\cite{lampert2009learning}) / 42.2(\cite{lampert2013attribute}) \\
%       ALE/HLE/AHLE \cite{akata2013label} & 37.4 / 39.0 / \textbf{43.5}  \\
%       Our method & 43.1 \\ \hline
%   \end{tabular}
% \end{wraptable}
label-embedding method \cite{akata2013label}, on the AwA
dataset.% with the same experimental setup.
We list the results of average multi-class classification accuracy in
Table~\ref{table:comparison_with_attribute}. Overall, compared to the
state-of-the-art attribute-based ZSL, our proposed method  achieves better or
comparable performance, especially compared to DAP and IAP. It should be
noted that all the attribute-based ZSL methods are based on the well-defined
visual attribute and the category-attribute relationship. In contrast, our
method does not depend on manually defined visual attributes; instead we only
exploit `free' semantic word space learned from linguistic knowledge bases
without the need for any manual annotation for the AwA classes. This is thus a
very encouraging result. If we apply the given visual attributes on AwA to do
the similarity computation, we can get 49.5\% performance, which is much
higher than the existing attribute-based methods.

\begin{table}
  \centering
  \label{table:comparison_with_attribute}
  \begin{tabular}{c|c|c}\hline
      Approach & semantic space & mean accuracy (in \%) \\ \hline
      DAP & attribute & 40.5(\cite{lampert2009learning}) / 41.4(\cite{lampert2013attribute}) \\
      IAP & attribute & 27.8(\cite{lampert2009learning}) / 42.2(\cite{lampert2013attribute}) \\
      ALE/HLE/AHLE \cite{akata2013label} & attribute & 37.4 / 39.0 / 43.5  \\
      Our method & word vector / attribute & 43.1 / \textbf{49.5} \\ \hline
  \end{tabular}
  \caption{Comparison with the state-of-the-art attribute-based zero-shot
    learning on AwA.}
\end{table}

\begin{figure}
  \centering
  \begin{minipage}{0.45\textwidth}
    \centering
    \includegraphics[width=0.9\linewidth]{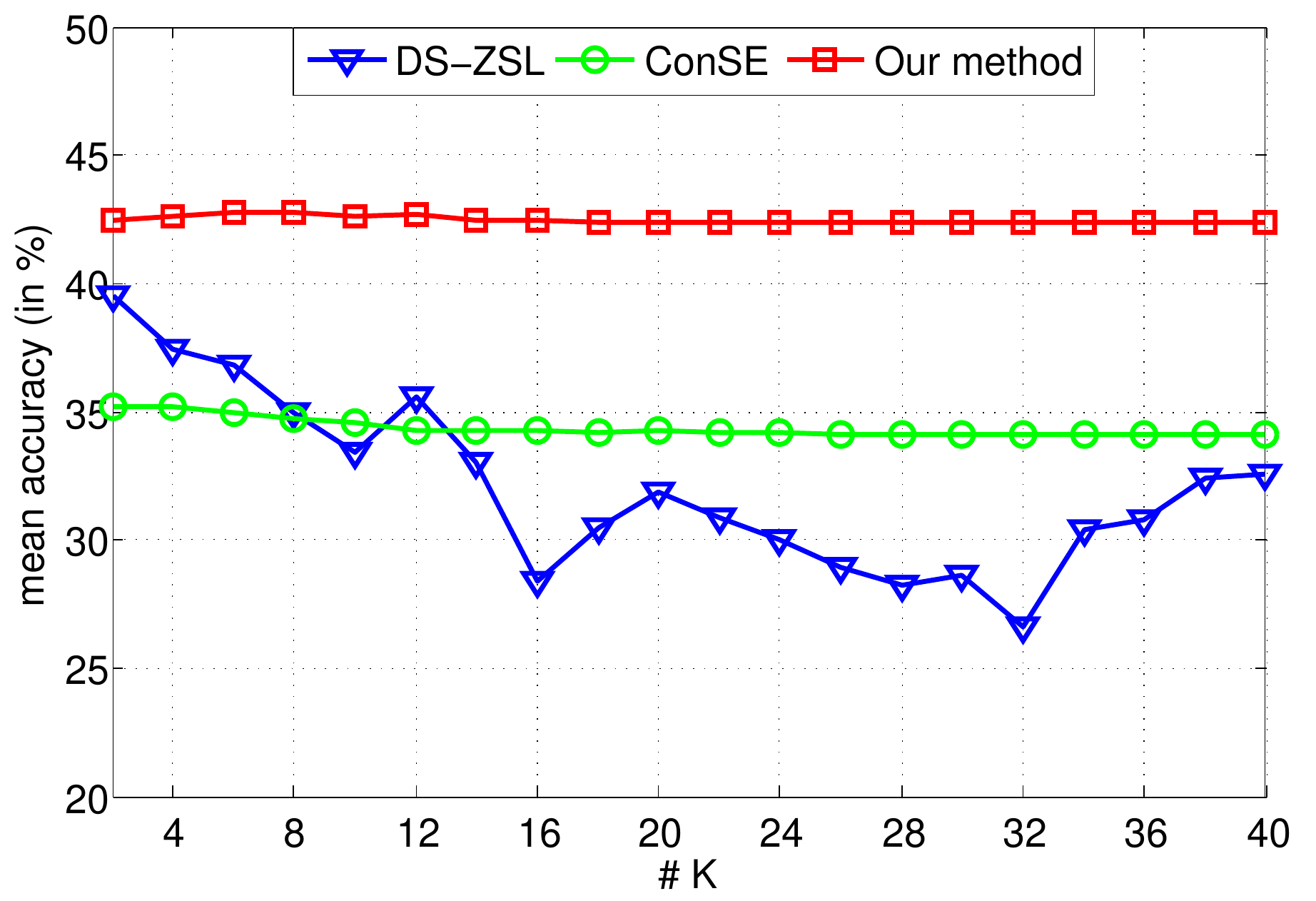}
    \caption{The performance (average multi-class classification accuracy in
      \%) of DS-ZSL, ConSE and our method with respect
    to different settings of the parameter $K$.}
    \label{fig:K_parameter}
  \end{minipage}
  \hspace{0.06\linewidth}
  \centering
  \begin{minipage}{0.45\textwidth}
    \centering
    \includegraphics[width=1\linewidth]{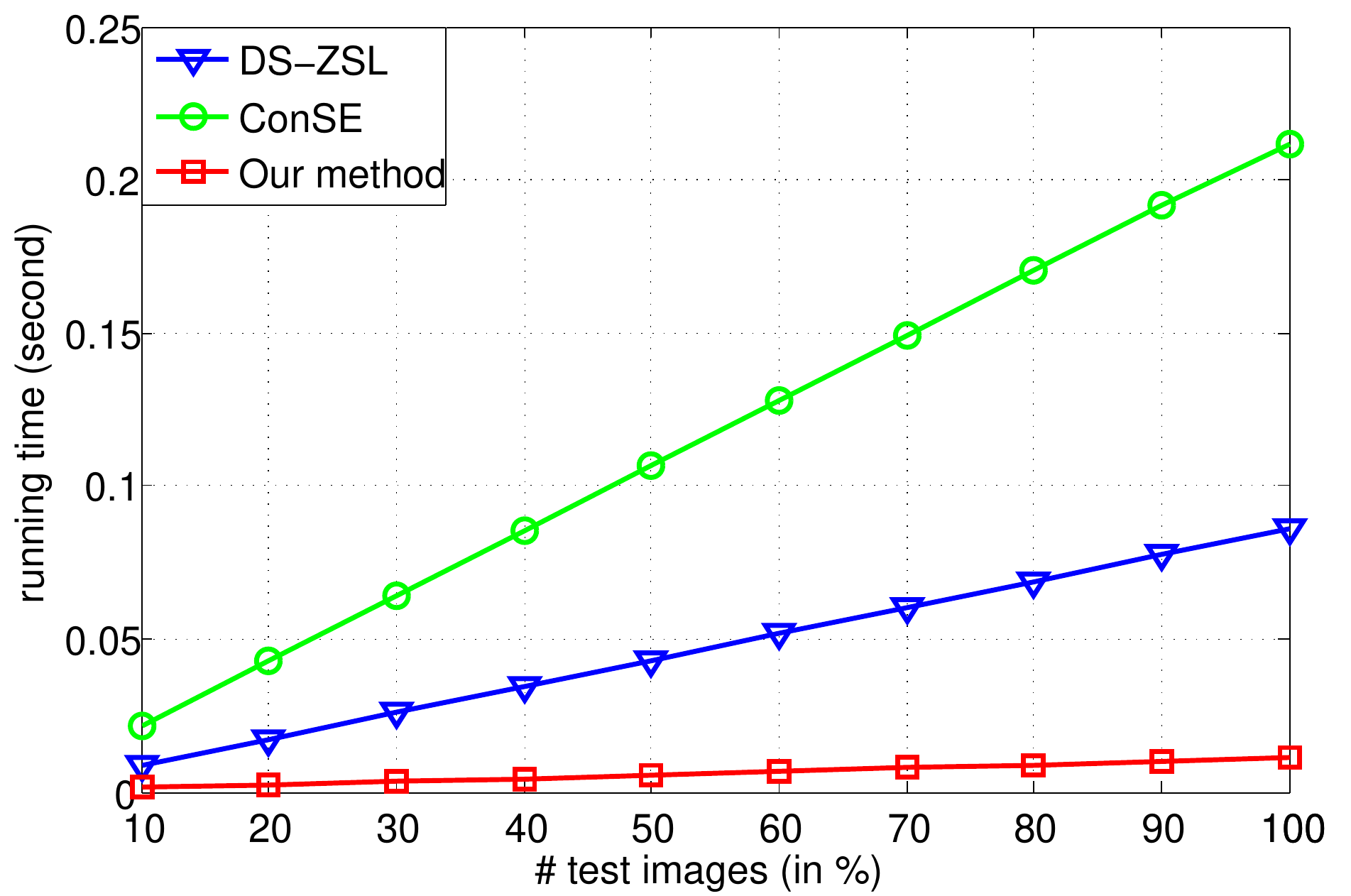}
    \caption{The running time of DS-ZSL, ConSE and our method with respect to
      different numbers of test images.}
    \label{fig:runtime}
  \end{minipage}
\end{figure}

\textbf{Parameter sensitivity.}
Since DS-ZSL, ConSE and our method have a same parameter $K$, i.e., the number
of top similar seen classes that a test image will choose, we analyze the
effect of setting different values of $K$ for the three methods. From
Fig.~\ref{fig:K_parameter}, we can see that DS-ZSL will be heavily affected by
the number of seen classes that connect with the test image, while ConSE and
our method are more stable. Especially, our method is almost not influenced by
the parameter $K$ at all. That is because through the more robust semantic
graph, our method can reduce the influence of the noisy seen classes which
will be inevitably included when the value of $K$ increases.

\textbf{Running time comparison.}
We also test the running time of DS-ZSL, ConSE and our method w.r.t. different
number of test images. There are totally 6180 test images on AwA. They are
divided into 10 folds and we test increasing number of folds of test images,
i.e. from 618 to 6180 and show the results in Fig.~\ref{fig:runtime}. We run
each algorithm 100 times at a PC machine with 3.9GHz and 16GB memory and
report the average result. From Fig.~\ref{fig:runtime}, we can see that all
the three methods are linear and our method is significantly faster than the
other two, especially given large number of test images.

%------------------------------------------------------------------------------------------------
\section{Conclusion}
\label{sec:conclusion}
In this work, we have introduced a novel zero-shot learning framework based on
semantic graph. The proposed method models the relationship among visual
categories using the semantic graph and then performs zero-shot learning
through an absorbing Markov chain process on the semantic graph. We have shown
experimentally that our method is more effective and more stable than the
alternative bipartite graph based methods.

%------------------------------------------------------------------------------------------------
{\small
\bibliographystyle{ieee}
\bibliography{nipsbib}
}

\end{document}